\documentclass[10pt,twocolumn,letterpaper]{article}

\usepackage{cvpr}              

\usepackage{amsthm}
\usepackage{amsmath}
\usepackage{wrapfig}  
\usepackage{tikz}
\usepackage{amssymb}
\usepackage{pifont}

\newcommand{\bs}[1]{\boldsymbol{#1}}

\definecolor{cvprblue}{rgb}{0.21,0.49,0.74}
\usepackage[pagebackref,breaklinks,colorlinks,allcolors=cvprblue]{hyperref}

\def\confName{CVPR}
\def\confYear{2026}

\usepackage{xspace}
\newcommand{\Reweaver}{\emph{ReWeaver}\xspace}

\title{ReWeaver: Towards Simulation-Ready and Topology-Accurate \\
Garment Reconstruction}

\author{
    Ming Li\textsuperscript{1,2,3}, 
    Hui Shan\textsuperscript{1,2,3}, 
    Kai Zheng\textsuperscript{3}, 
    Chentao Shen\textsuperscript{1,3},\\ 
    Siyu Liu\textsuperscript{6}, 
    Yanwei Fu\textsuperscript{2,4}, 
    Zhen Chen\textsuperscript{5}, 
    Xiangru Huang\textsuperscript{3} \\
    \and
    \textsuperscript{1}Zhejiang University \quad \textsuperscript{2}Shanghai Innovation Institute
    \quad \textsuperscript{3}Westlake University\\
    \quad \textsuperscript{4}Fudan University
    \quad \textsuperscript{5}Adobe
    \quad \textsuperscript{6}Xidian University
    \\
}

\begin{document}
\twocolumn[{%
\renewcommand\twocolumn[1][]{#1}%
\maketitle

\begin{center}

\centering
    \begin{tikzpicture}
    \node[anchor=south west,inner sep=0] (image) at (0,0){\includegraphics[width=\linewidth]{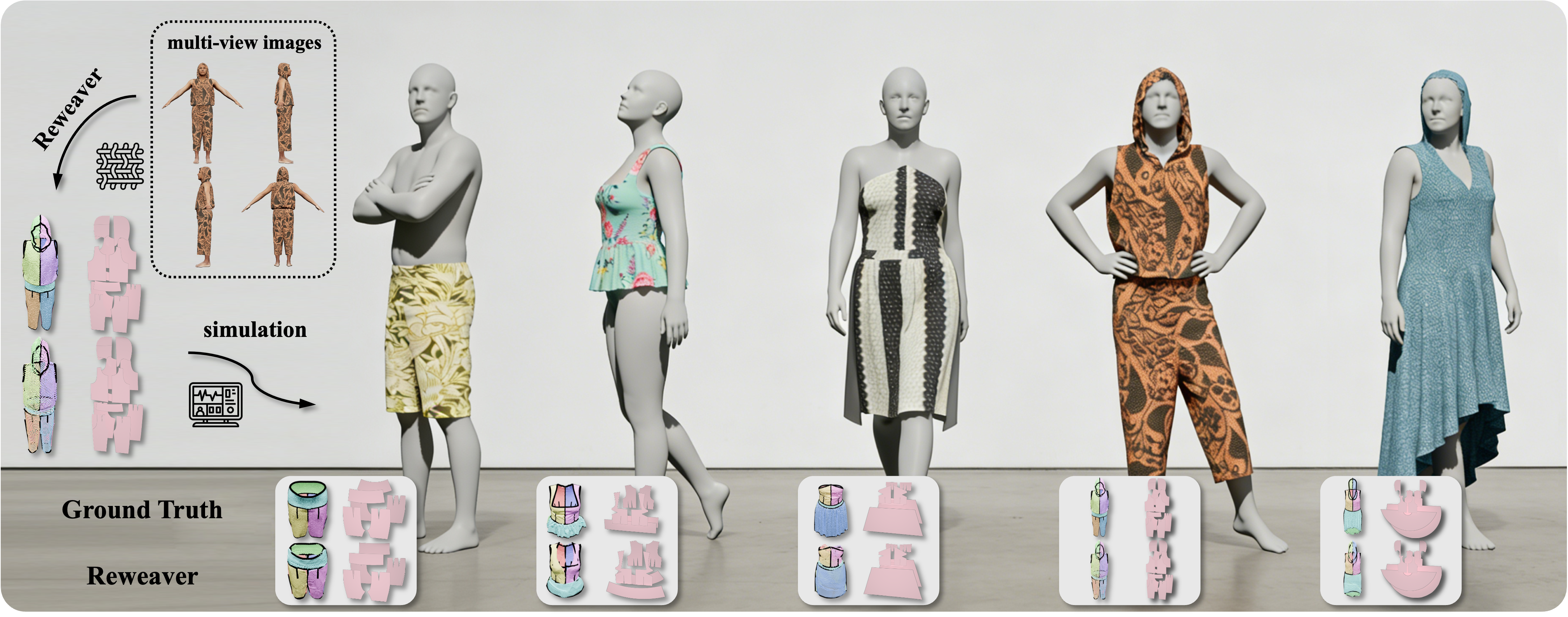}};
    \begin{scope}[x={(image.south east)},y={(image.north west)}]
    \end{scope}
    \end{tikzpicture}
    \captionof{figure}{\textbf{ReWeaver.} From as few as four input views, \Reweaver reconstructs high-precision sewing patterns with complex topology together with their corresponding 3D geometry. The method outputs a unified 2D--3D garment representation, where each panel and edge is explicitly linked to its associated 3D points. This enables faithful, simulation-ready garment assets to be recovered from ordinary and sparse-view photographs without a controlled capture setup. The reconstructed garment geometry and topology is precisely aligned with the input images and can be used for 3D structural perception.}
\end{center}
}]


\begin{abstract}
High-quality 3D garment reconstruction plays a crucial role in mitigating the sim-to-real gap in applications such as digital avatars, virtual try-on and robotic manipulation. However, existing garment reconstruction methods typically rely on unstructured representations, such as 3D Gaussian Splats, struggling to provide accurate reconstructions of garment topology and sewing structures. As a result, the reconstructed outputs are often unsuitable for high-fidelity physical simulation. We propose \textbf{ReWeaver}, a novel framework for topology-accurate 3D garment and sewing pattern reconstruction from \textit{sparse} multi-view RGB images. Given as few as four input views, ReWeaver predicts seams and panels as well as their connectivities in both the 2D UV space and the 3D space. The predicted seams and panels align precisely with the multi-view images, yielding structured 2D--3D garment representations suitable for 3D perception, high-fidelity physical simulation, and robotic manipulation.
To enable effective training, we construct a large-scale dataset~\textbf{GCD-TS}, comprising multi-view RGB images, 3D garment geometries, textured human body meshes and annotated sewing patterns. The dataset contains over 100,000 synthetic samples covering a wide range of complex geometries and topologies. Extensive experiments show that ReWeaver consistently outperforms existing methods in terms of topology accuracy, geometry alignment and seam-panel consistency. Code and data will be available at the project \href{https://sii-liming.github.io/ReWeaver/}{sii-liming.github.io/ReWeaver/}.

High-quality 3D garment reconstruction plays a crucial role in mitigating the sim-to-real gap in applications such as digital avatars, virtual try-on and robotic manipulation. However, existing garment reconstruction methods typically rely on unstructured representations, such as 3D Gaussian Splats, struggling to provide accurate reconstructions of garment topology and sewing structures. As a result, the reconstructed outputs are often unsuitable for high-fidelity physical simulation. We propose ReWeaver, a novel framework for topology-accurate 3D garment and sewing pattern reconstruction from sparse multi-view RGB images. Given as few as four input views, ReWeaver predicts seams and panels as well as their connectivities in both the 2D UV space and the 3D space. The predicted seams and panels align precisely with the multi-view images, yielding structured 2D-3D garment representations suitable for 3D perception, high-fidelity physical simulation, and robotic manipulation.To enable effective training, we construct a large-scale dataset GCD-TS, comprising multi-view RGB images, 3D garment geometries, textured human body meshes and annotated sewing patterns. The dataset contains over 100,000 synthetic samples covering a wide range of complex geometries and topologies. Extensive experiments show that ReWeaver consistently outperforms existing methods in terms of topology accuracy, geometry alignment and seam-panel consistency. Code and data will be available at the project sii-liming.github.io/ReWeaver.
\end{abstract}    
\section{Introduction}
\label{sec:intro}

High-quality 3D garment reconstruction is central to many emerging applications in virtual try-on, digital humans, gaming, and robotic manipulation~\cite{garment_important,cloth4d,cloth3d,garment_important_2}. For these tasks, it is not sufficient to recover only a visually plausible garment surface: downstream simulation, animation, and asset creation require \emph{structured} representations that capture how garments are constructed—namely, their sewing patterns, panels, and seams.

Existing reconstruction methods typically operate on unstructured representations such as point clouds, noisy meshes, unsigned or signed distance fields, or 3D Gaussian splats~\cite{gaussian_garment,garment_recon_1,garment_recon_2,garment_recon_3,garment_recon_4,garment_recon_5,garment_recon_6}. While these formats can approximate garment geometry, they lack explicit sewing structure, making them difficult to parameterize for physics-based simulation, garment editing, or retargeting. Moreover, these representations are inherently misaligned with industry-standard garment design workflows, which rely on 2D sewing patterns as the primary medium for design and manufacturing.

Motivated by these limitations, recent works have shifted toward learning sewing pattern structures directly~\cite{diffavatar,dress-1-to-3,dresscode,chatgarment,aipparel,physavatar,sewformer,neuraltailor,diffusion_gen_2,sig_garment_image}. However, these approaches have significant shortcomings. Methods relying on predefined topologies~\cite{diffavatar} perform well only on simple garments and fail to handle unseen layouts. Vision–language-model–based approaches~\cite{chatgarment,aipparel} predict tokenized JSON descriptions that are later converted into 2D patterns with GarmentCode~\cite{garmentcode} framework. While this approach generalizes to a broader range of topologies, the geometric accuracy of the reconstruction remains unsatisfactory.

To simultaneously reconstruct accurate \emph{topology} and \emph{geometry} of 3D garment, we propose \Reweaver, 
 a unified model that reconstructs both 3D garment structure and 2D sewing patterns from as few as four input views. \Reweaver predicts a set of 3D patches and curves together with their 2D counterparts, establishing explicit 2D–3D correspondences. The predicted 3D patches and curves align closely with the multi-view images, enabling accurate 3D perception and robotic tasks. At the same time, the predicted sewing patterns further enable \Reweaver to extract simulation-ready and topology-accurate 3D garment assets from in-the-wild sparse view images.


To accommodate for sparse and unknown distribution of input views, we adopt the multi-view visual encoder from VGGT~\cite{vggt}. The visual encoder alternates between intra-frame attention and inter-frame attention and outputs a set of encoded tokens. We reconstruct the 3D garment geometry and topology by predicting a set of 3D curves, a set of 3D patches, and the patch-curve connectivity. Using the predicted patch-curve connectivity, we group each 3D patch tokens with its connecting curves tokens and perform intra-group attention to flatten the 3D curves into 2D panel edges.
 



Compared to existing methods, \Reweaver achieves superior garment reconstruction quality and, for the first time, offers garment geometry and topology reconstruction in both 2D and 3D spaces, with precise alignment to the input images. This allows \Reweaver to be employed in various downstream applications such as 3D structural perception, physical simulation, and garment asset creation. 

Our main contributions can be summarized as follows:
\begin{enumerate}
    \item We introduce \textbf{ReWeaver}, the \emph{first} framework to jointly reconstruct structured 3D garments and 2D sewing patterns while maintaining 2D–3D correspondence.
    \item We present \textbf{GCD-TS} (GarmentCodeData with extended textures and seam annotations), an extension of the large-scale GCD dataset that adds garment and body texture assets and provides structured 3D point-cloud annotations, together with their explicit correspondences to 2D panels.
    \item Our model outperforms prior methods in pattern reconstruction, achieving strong topological generalization while maintaining high geometric fidelity to the input images.
\end{enumerate}
\section{Related Works}
\label{sec:related_work}

\subsection{Learning-based garment reconstruction and generation}
With the advancement of deep learning techniques and the construction of large-scale garment datasets~\cite{garmentcodedata,garment_pattern_generator,cloth3d,sewformer,chatgarment,aipparel}, a variety of data-driven methods have been proposed for garment pattern reconstruction. NeuralTailor~\cite{neuraltailor} is a pioneering work that focuses on learning-based recovery of structured garment sewing patterns from 3D point clouds, and demonstrates generalization to some unseen garment types. Going further, SewFormer~\cite{sewformer} introduces a two-level transformer network to directly predict sewing patterns from a single image. With the rapid development of vision-language models (VLMs)~\cite{llava,gpt4v,gpt4o}, recent works~\cite{aipparel,dresscode,chatgarment} have leveraged VLMs for data annotation or fine-tuning, enabling text-conditioned and image-conditioned sewing pattern reconstruction, generation, and editing. DressCode~\cite{dresscode} presents the first text-driven garment generation pipeline that produces high-quality sewing patterns and physically-based textures, demonstrating the feasibility of using a GPT-based autoregressive model for sewing pattern generation tasks. \cite{chatgarment,aipparel} further empower the model to generate sewing patterns with different tokenization schemes by fine-tuning LLaVA~\cite{llava}. Recent works explore diffusion-based denoising in a latent space and then decode to various pattern representations~\cite{garmagenet,diffusion_gen_1,diffusion_gen_2}. With the exception of~\cite{garmagenet}, these methods focus exclusively on 2D patterns and largely overlook precise 3D geometric understanding.

\subsection{Optimization-based garment reconstruction}
Some optimization-based methods, as well as hybrid approaches that combine learning-based prediction with physics-guided optimization, have made notable progress in recovering simulation-ready garments by integrating differentiable physics into the reconstruction process. 
For example, DiffAvatar~\cite{diffavatar} leverages the XPBD~\cite{xpbd} algorithm to drive a differentiable physics engine that simulates cloth behavior, allowing gradient-based updates to align the simulated garment with the reconstructed scan geometry. However, this method relies on a predefined garment library, which limits its ability to generalize to diverse or unseen pattern structures. Furthermore, optimizing the pattern parameters requires running the differentiable simulation multiple times, making the overall process highly time-consuming.
Dress-1-to-3~\cite{dress-1-to-3} addresses a more challenging setting by leveraging generative models as priors and using the C-IPC~\cite{cipc} simulation algorithm to reconstruct simulation-ready garments from a single input image. Its initial garment patterns are estimated from a single image using SewFormer~\cite{sewformer}, but these patterns have low geometric accuracy and may even contain incorrect topology. 
Gaussian Garments~\cite{gaussiangarment} reconstructs an initial static geometry from a single template frame via multi-view stereo, surface reconstruction, and remeshing, and subsequently registers it to the video sequence. 
However, the acquisition of the static geometry relies on multiple traditional steps, suffering from error accumulation, and such reconstruction requires the selected template frame to possess high visibility and rich textures.

\section{Method}
\label{sec:method}

We introduce \textbf{\Reweaver}, a framework that reconstructs accurate geometry and topology of 3D garments from multi-view images with as few as four views. In the rest of this section, we first formally define the garment reconstruction task in Section~\ref{subsec:problem:setup}, then introduce the building blocks of \Reweaver: a multi-view visual encoder adopted from VGGT~\cite{vggt} (Section~\ref{subsec:visual:encoder}); a consecutive module that predicts 3D curve and patch using a bi-path transformer (Section~\ref{subsec:curve:patch:prediction}); and an intra-surface attention module that implicitly learns a flattening process that converts the 3D curves and patches into their 2D counterparts (Section~\ref{subsec:flattening}). Finally, we describe the loss functions used for training in Section~\ref{subsec:loss:function}.

\begin{figure}[t]
  \centering
  \includegraphics[trim={70pt 80pt 70pt 80pt}, width=\linewidth]{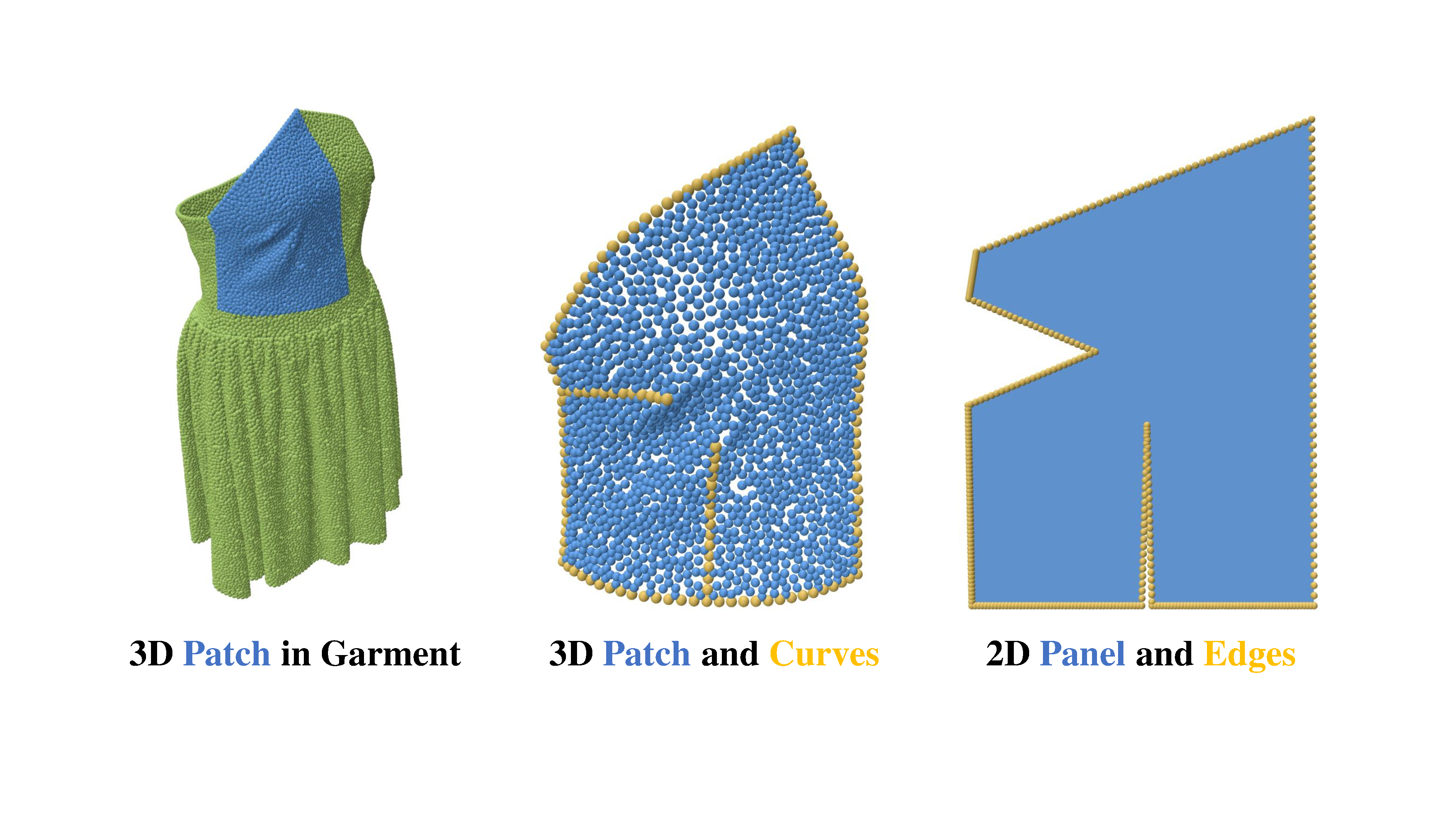}
  \vspace{-0.5cm}
  \caption{Visualization of the terminologies used in this paper.}
  \label{fig:show_term}
\end{figure}

\textbf{Terminology.} Since we operate in both 2D and 3D spaces, we adopt the following terminology for clarity. We denote 3D garment surface regions as \emph{patches} and their 2D counterparts as \emph{panels}. We use \emph{curve} to refer to the 3D locus of a seam or boundary, and \emph{edge} to denote its 2D counterpart on the panel. See Figure~\ref{fig:show_term} for a visualization.

\begin{figure*}[t]
    \centering
    \includegraphics[width=\textwidth]{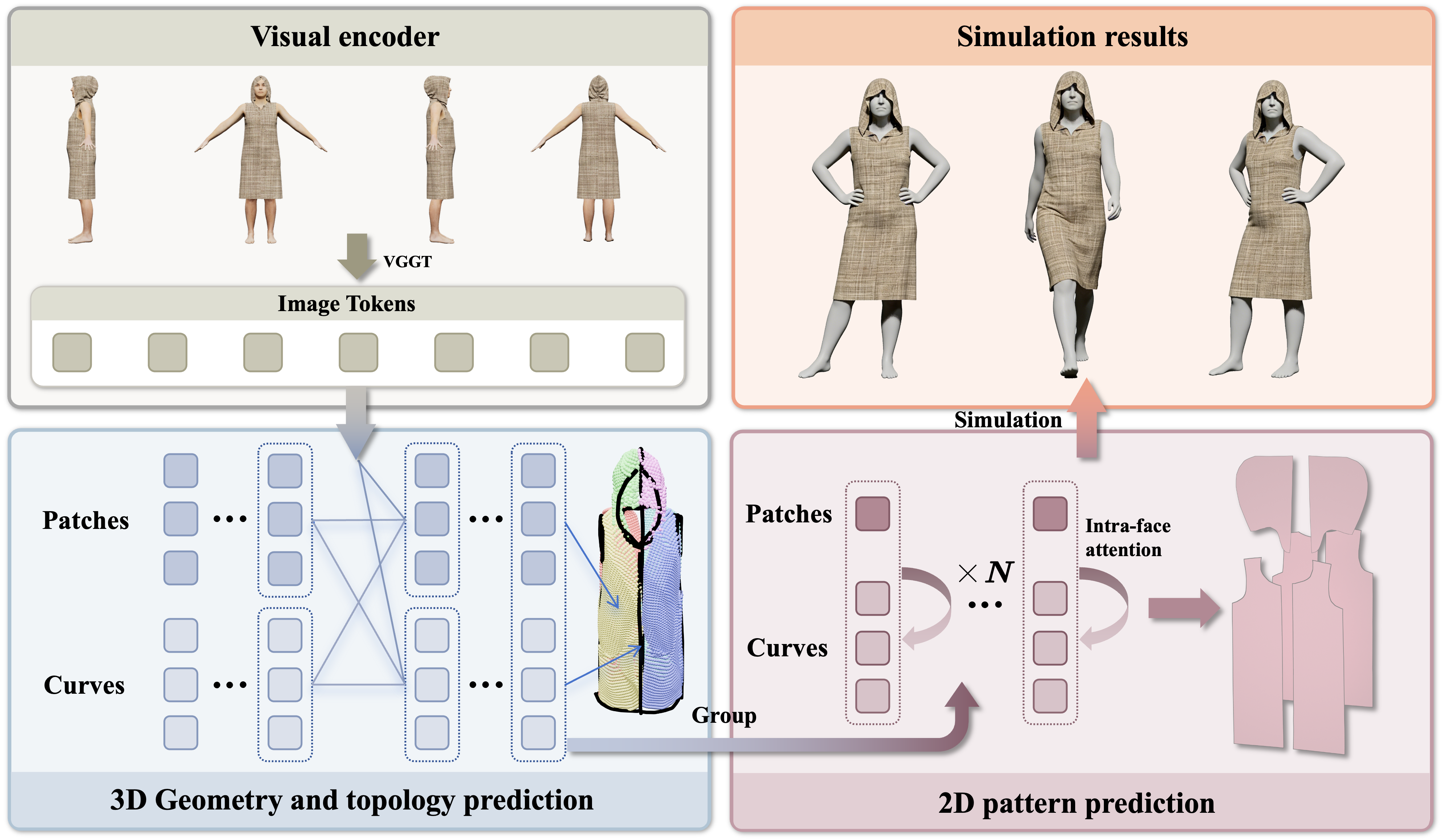}
    \caption{\textbf{Pipeline of our method.} Our VGGT-like image encoder extracts features from multi-view images (Section  \ref{subsec:visual:encoder}), which then interact with predefined patch and curve queries. In the 3D geometry and topology prediction module (Section~\ref{subsec:curve:patch:prediction}), these queries and the image tokens pass through stacked self- and cross-attention blocks. The resulting tokens are decoded into 3D curves and patches. The same tokens are then reused in the 2D pattern prediction module (Section~\ref{subsec:flattening}): guided by the refined topology, we group the patches and curves into patch-centric groups, apply intra-group attention, and finally decode the edges of the 2D panels. The decoded 2D panels can be directly used for physical simulation.}
    \label{fig:pipeline}
\end{figure*}

\subsection{Problem Setup}
\label{subsec:problem:setup}

\textbf{\Reweaver} addresses the task of garment reconstruction from multi-view RGB images. 
Given a set of images 
$\mathcal{I} = \{ I_j \mid I_j \in \mathbb{R}^{H \times W \times 3} \}$ 
capturing a 3D garment, the goal is to reconstruct \emph{four} groups of elements:

\begin{enumerate}[leftmargin=*]
  \item \textbf{3D curves:} 
  $\mathcal{C} = \{ C_j \mid C_j \in \mathbb{R}^{M_C \times 3} \}$, 
  representing the garment seams in 3D space.

  \item \textbf{3D patches:} 
  $\mathcal{P} = \{ P_i \mid P_i \in \mathbb{R}^{M_P \times 3} \}$, 
  representing garment surfaces in 3D space, each 3D patch is topologically equivalent to a 2D square.

  \item \textbf{2D edges:} 
  $\mathcal{E} = \{ E_{ij} \mid E_{ij}\in\mathbb{R}^{M_E\times 2} \}$, where for patch $P_i$ and its attached 3D curves $C_j$, $E_{ij}$
  denotes the 2D flattened boundary edge corresponding to $C_j$. \footnote{
  A single 3D seam $C_j$ can induce multiple 2D edges across its incident patches. Thus, we employ two indices to denote each individual edge.
  }

  \item \textbf{Patch–curve connectivity:} 
  A binary matrix 
  $\sigma_{pc} \in \{0,1\}^{N_P \times N_C}$ 
  where $\sigma_{pc}(i, j) = 1$ indicates patch $i$ is attached to curve $j$.
  
\end{enumerate}

Here, $N_P$ and $N_C$ denote the number of patches and curves, respectively, while $M_P$, $M_C$, and $M_E$ represent the number of points used to model each patch, curve, and edge. Together, these elements provide a unified multi-level representation. 
The 3D patches and curves support tasks such as \emph{3D structural perception} and shape analysis, 
while the 2D edges and patch–curve connectivity enable \emph{2D asset extraction} for garment pattern design and physics-based simulation (via 2D triangulation of panel interiors bounded by $\{E_{ij}\}$).

\subsection{Visual Encoder}
\label{subsec:visual:encoder}
We adopt a multi-view visual encoder design following VGGT~\cite{vggt}. Each input image is divided into non-overlapping patches that are embedded into tokens using a DINOv2 backbone~\cite{dinov2}. 
Tokens are processed by alternating intra-frame and inter-frame self-attention layers:
the intra-frame layers refine per-view features, while the inter-frame layers aggregate information across views. Following VGGT~\cite{vggt}, these attention blocks are alternately stacked, progressively integrating local and global geometric cues into unified token representations.
We concatenate the outputs of the final inter-and intra-frame layers and flatten all frame tokens into a sequence
$T_i \in \mathbb{R}^{N_i\times D}$,
where $D$ is the embedding dimension.


\subsection{3D Curve and Patch Prediction}
\label{subsec:curve:patch:prediction}
Given the token sequence $T_i$ processed by the visual encoder, we design a topology and geometry prediction module that predicts the 3D curves, patches, and their connectivities. Following existing works in 3D object detection from multi-view images~\cite{query_design,wang2022detr3d}, the prediction module
uses latent patch and curve queries together with attention to make direct 3D predictions and support diverse garment topologies with an undetermined number of curves and patches. 

Specifically, the prediction module takes as input the token sequence $T_i$ and a set of learnable patch queries $Q_p \in \mathbb{R}^{N_p \times D}$ and curve queries $Q_c \in \mathbb{R}^{N_c \times D}$, and output three groups of elements: 1) the probability of each queried patch or curve being valid; 2) the predicted geometry of each curve and patch; and 3) the predicted patch-curve connectivity. 
We next detail the overall architecture design and the decoding heads used for different prediction tasks.


\paragraph{Processing patch and curve queries.} 
Inspired by ComplexGen~\cite{complexgen}, we adopt a bi-path Transformer architecture that models interactions both between patch and curve tokens and between these tokens and the image tokens. At each layer of the attention module, an element group (patch or curve) first performs self-attention to conduct intra-group message passing, and then applies cross-attention to retrieve contextual information from both the image tokens and the other element group. This design enables full fusion of image evidence while maintaining global structural and geometric consistency. 
After these attention operations, each path is passed through a layer normalization~\cite{layernorm} and a feed-forward module, followed by a residual connection~\cite{resnet}. After the final layer, we obtain the refined tokens 
$T_p \in \mathbb{R}^{N_p \times D}$ 
and 
$T_c \in \mathbb{R}^{N_c \times D}$, 
which serve as the processed representations of the patch and curve queries, respectively. 

\paragraph{Probability Prediction.} 
The probability decoding heads separately decode tokens $T_p$ and $T_c$ into probability vectors for patches and curves, denoted as $\sigma_p \in [0, 1]^{N_p}$ and $\sigma_c \in [0, 1]^{N_c}$, respectively. For patches and curves, we use two separate decoding heads $f_p^{\textup{prob}}$ and $f_c^{\textup{prob}}$, each consisting of a three-layer MLP followed by a sigmoid to normalize outputs to $[0, 1]$:
\begin{equation*}
    \sigma_p^i = \textup{sigmoid}\big(f_p^{\textup{prob}}(T_p^i)\big), 
    \quad
    \sigma_c^i = \textup{sigmoid}\big(f_c^{\textup{prob}}(T_c^i)\big).
\end{equation*}
Low-probability elements are filtered by thresholds $\epsilon_p$ and $\epsilon_c$, and the topology is further refined (Appendix~\ref{app:topo:refine}) to obtain binary validity masks 
$\bs{\sigma}_p^{\star}\in\{0, 1\}^{N_p}$ and $\bs{\sigma}_c^{\star}\in\{0, 1\}^{N_c}$.

\paragraph{Geometry Prediction.}
We represent each patch and curve as a continuous parametric function that maps canonical coordinates to 3D space. 
Specifically, these functions are implemented as three-layer MLPs that map $[0,1]$ and $[0,1]^2$ domains to $\mathbb{R}^3$. 
Inspired by the hyper-network formulations~\cite{hyper_net1,hyper_net2}, we define the geometry prediction heads 
$f_c^{\textup{geo}}$ and $f_p^{\textup{geo}}$ as hyper-networks that generate the weights of these MLPs, conditioned on the processed query tokens 
$T_c$ and $T_p$. 
Each curve or patch token thus parameterizes a unique mapping:
\begin{equation}
\begin{aligned}
    \forall u \in [0, 1],&~\quad g_c^i(u) = f_c^{\textup{geo}}(T_c^i)(u) \in \mathbb{R}^3\\
    \forall u,v \in [0,1],&~\quad g_p^i(u,v) = f_p^{\textup{geo}}(T_p^i)(u,v) \in \mathbb{R}^3,
\end{aligned}
\end{equation}
where $g_c^i$ and $g_p^i$ denote the instantiated MLPs that generate 3D points along the $i$-th curve and patch, respectively.

During training, we uniformly sample $u \in [0,1]$ and $(u,v) \in [0,1]^2$,
apply the corresponding MLPs to obtain the predicted point sets, and supervise them using point-based geometric losses. This implicit formulation enables differentiable supervision at arbitrary sampling densities without degrading geometric smoothness or continuity.


\paragraph{Connectivity Prediction.} 
To predict the connectivity between patch $i$ and curve $j$, we project the patch and curve tokens with two linear layers $f^{\textup{adj}}_p$ and $f^{\textup{adj}}_c$, take their dot product, and apply a sigmoid to map it to $[0,1]$ as the adjacency probability, i.e.,
\begin{equation}
    \sigma_{pc}(i, j) = \textup{sigmoid}(f^{\textup{adj}}_p(T_p^i) \cdot f^{\textup{adj}}_c(T_c^j))
\end{equation}
adjacency All these probabilities form an adjacency matrix $\sigma_{pc}$, which is first preliminarily filtered by the threshold $\epsilon_{\textup{adj}}$ and then refined via a topology refinement procedure (Appendix~\ref{app:topo:refine}). The refined adjacency matrix is denoted as $\sigma^{\star}_{pc} \in \{0,1\}^{N_p\times N_c}$.

\subsection{2D Pattern Prediction}
\label{subsec:flattening}
Given the valid patch and curve tokens ($T_p$, $T_c$) and their refined topology ($\bs{\sigma}_p^{\star}$, $\bs{\sigma}_c^{\star}$, $\sigma_{pc}^{\star}$), 
we group each valid patch token with its connected curve tokens. 
Each group is processed by an intra-group attention module, resulting in edge tokens $T_e$ that are then decoded into 2D panel edges.

Specifically, for each group, the curve tokens are first processed via self-attention to exchange information, followed by cross-attention with the associated patch token. We then apply the same layer normalization, feed-forward module, and residual connection as in Section ~\ref{subsec:curve:patch:prediction}. The resulting edge tokens are denoted as $T_e$. Let $\partial_i = \{\,j | \sigma_{pc}^{\star}(i,j)=1\,\}$ denote the indices of curve tokens connected to patch $i$, and let $T_e^{\partial_i}$ denote the corresponding edge tokens. 
For a connected curve $j\in \partial_i$, we use a three-layer hyper-network to generate an MLP that maps a canonical 1D parameter to normalized 2D coordinates:
\begin{equation}
\forall u\in[0,1],~\quad g_e^{ij}(u) = f_e^{\textup{edge}}(T_e^{j})(u) \in [0, 1]^2.
\end{equation}
Here $g_e^{ij}$ denotes the instantiated edge function that produces the 2D edge corresponding to the $j$-th curve connected to patch $i$, and $f_e^{\textup{edge}}$ denotes another hyper-network that takes in the edge token and output an MLP, similar to other previously defined hyper-networks. Because this formulation does not guarantee that the endpoints of adjacent edges meet perfectly, we apply a geometric refinement procedure (Appendix~\ref{app:geo:refine}) to enforce loop closure with a high success rate.


Since each 2D panel is predicted in a normalized coordinate system within $[0,1]^2$, to recover the absolute panel metrics, we introduce an additional lightweight MLP $f_p^{\textup{scale}}$ that predicts a scalar scale factor $s_i$
from the corresponding patch token $T_p^i$:
\begin{equation}
s_i = f_p^{\textup{scale}}(T_p^i), \quad s_i \in \mathbb{R}.
\end{equation}
The final scaled 2D panel is obtained by multiplying its normalized coordinates by $s_i$. 
This scale term allows the model to represent panels at realistic physical sizes and supports accurate downstream garment reconstruction.

\subsection{Loss Functions}
\label{subsec:loss:function}

In this section, we elaborate how to compute differentiable losses for supervised learning. 
We first establish the correspondence between the predicted elements and the ground-truth annotations 
(patches, curves) using the Hungarian matching algorithm~\cite{hungarian_matching,hungarian_matching_2}.

\paragraph{Geometric Loss.}
For all parametric functions predicted by the hyper-networks (patches, curves, and edges), 
we supervise their output geometries using the Chamfer Distance (CD) loss. 
Let $\mathcal{G} = \{ g_p^i, g_c^j, g_e^{ij} \}$ denote the set of all predicted geometric mappings, 
and $m(g)$ be the ground-truth element matched to $g$ through Hungarian matching. 
The geometric loss is computed over sampled point sets from each predicted mapping:
\begin{equation}
L_{\textup{geo}} =
\sum_{g \in \mathcal{G}}
w_{\textup{geo}}^{(g)} \,
\textup{CD}\big(V(g), V(m(g))\big),
\end{equation}
where $V(g)$ and $V(m(g))$ are the sampled 2D or 3D point sets from the predicted and ground-truth geometries, respectively.

\paragraph{Classification and Connectivity Loss.}
For the predicted validity and connectivity probabilities 
$(\bs{\sigma}_p, \bs{\sigma}_c, \sigma_{pc})$, 
we apply Binary Cross-Entropy (BCE) supervision:
\begin{equation}
L_{\textup{cls}} = 
\sum_{\sigma \in \{\bs{\sigma}_p, \bs{\sigma}_c, \sigma_{pc}\}} 
w_{\textup{cls}}^{(\sigma)} \,
\textup{BCE}\big(\sigma, m(\sigma)\big),
\end{equation}
where $m(\sigma)$ denotes the corresponding ground-truth label.

\paragraph{Scale Loss.}
For the 2D panel scale predictions, we use an $\ell_2$ loss:
\begin{equation}
L_{\textup{scale}} =
\sum_{i=1}^{N_p}
w_{\textup{scale}} \,
\| s_i - s_{m(i)}^{\textup{gt}} \|_2^2.
\end{equation}

\paragraph{Total Loss.}
The overall training objective combines all above loss functions, 
where the weighting coefficients $w_{\textup{cls}}^{(\cdot)}$, $w_{\textup{geo}}^{(\cdot)}$, and $w_{\textup{scale}}$ 
are empirically determined (see Appendix~\ref{app:exp:details} for details).





\section{Experiments}
\label{sec:experiments}

\subsection{GCD-TS Dataset}
\label{subsec:GCD-TS}
\begin{figure}[htbp]
    \centering
    \begin{tikzpicture}
        \node[anchor=south west,inner sep=0] (image) at (0,0){\includegraphics[width=0.8\linewidth]{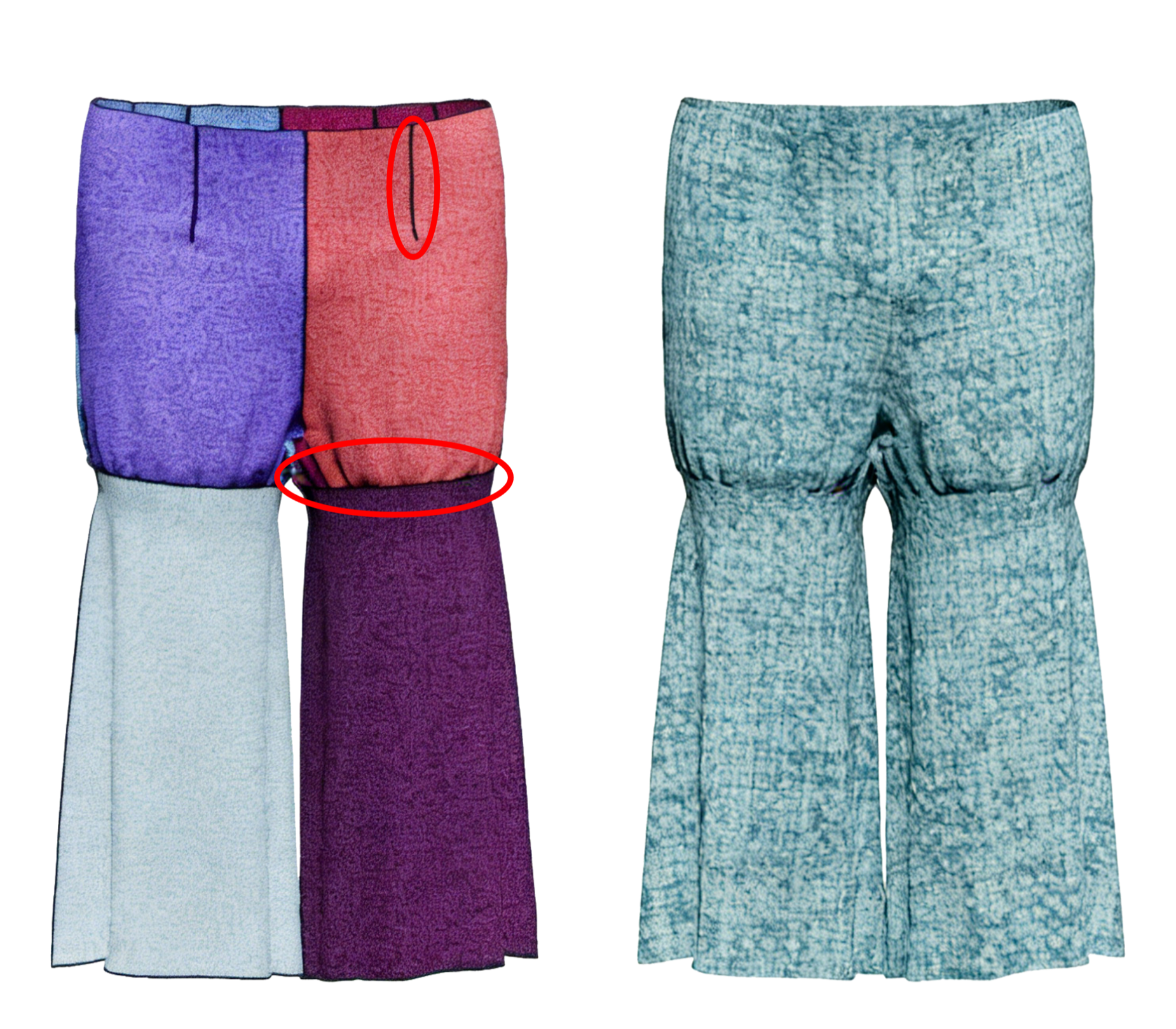}};
        \begin{scope}[x={(image.south east)},y={(image.north west)}]
            \node[anchor=south] at (0.25, 1) {\textbf{GCD}};
            \node[anchor=south] at (0.75, 1) {\textbf{GCD-TS}};
        \end{scope}
    \end{tikzpicture}
    
    \vspace{-0.2cm}
    
    \caption{\textbf{Texture differences between GCD and GCD-TS.} Using the same garment geometry, GCD textures reveal strong seam cues (e.g., highlighted regions), which are unrealistic and can lead to overfitting. GCD-TS replaces these with more realistic, diverse textures to improve generalization.}
    \label{fig:texture-difference}
  \end{figure}

To construct the GCD-TS dataset, we follow the garment sampling and simulation procedure of GCD~\cite{garmentcodedata,garmentcode} using both male and female SMPL~\cite{loper2023smpl} bodies. To improve realism and generalization, we replace the original GCD textures—which contain strong seam cues and textureless bodies—with a diverse set of garment and body textures. For body textures, we adopt nearly 50 textures from BEDLAM~\cite{bedlam}.  
For garment textures, we randomly sample from our collected library of tileable textures drawn from open-source sources~\cite{texweb1} and commercial datasets~\cite{texweb2}. Each garment–body pairing is rendered from four viewpoints (front, back, left, right), with small randomized perturbations in camera pose. In total, GCD-TS contains roughly \textbf{100,000} textured multi-view samples. Figure~\ref{fig:texture-difference} illustrates the difference between GCD and GCD-TS textures.

\subsection{Experimental Setup}
\label{subsec:exp:setup}

\textbf{Training Details.} We randomly split the GCD-TS dataset into training, validation, and test sets with a ratio of 8:1:1, and trained our model on the training split. We adopted the lightweight 21 MB pre-trained weights of DINOv2~\cite{dinov2} to initialize the image feature extraction module. The numbers of patch and curve queries were set to 200 and 70, respectively --- approximately twice the maximum counts observed in the training data. Each input sample consists of four RGB images with a resolution of $518\times518$. Each curve and its corresponding edge are represented by 50 uniformly sampled points, and each patch is represented by a $20\times20$ point set. All tokens, including image, curve, and patch tokens, have dimension 768. See Appendix~\ref{app:exp:details} for more implementation and training details.

\textbf{Evaluation Metrics.} To quantitatively evaluate our sewing pattern predictions, we define a set of evaluation metrics and also adopt several indicators from previous works~\cite{aipparel,garmagenet}. Before computing the 2D geometric metrics, we zero-center each panel. To assess the accuracy of topological prediction, we report: 1) \textbf{Panel Count Accuracy ($\text{Acc}_p$)}, the percentage of sewing patterns with a correctly predicted number of panels; 2) \textbf{Edge Count Accuracy ($\text{Acc}_e$)}, the percentage of correctly predicted edge counts within each correctly predicted panel; and 3) \textbf{Overall Accuracy ($\text{Acc}_o$)}, defined as the product of Panel Count Accuracy and Edge Count Accuracy, providing a comprehensive measure of garment topology reconstruction quality. To evaluate 2D geometric precision, we report: 4) \textbf{2D Edge CD ($\text{CD}_e$)}, the Chamfer Distance between the closed-loop edge point sets of the predicted and ground-truth panels.; and 5) \textbf{2D Panel IoU (IoU)}, the pixel-wise Intersection over Union between rasterized images of the predicted and ground-truth panels. To evaluate 3D geometric accuracy, we report: 6) \textbf{Patch Point Cloud CD ($\text{CD}_p$)}, the Chamfer Distance between the predicted and ground-truth 3D patch point sets; and 7) \textbf{3D Curve CD ($\text{CD}_c$)}, the Chamfer Distance between the predicted and ground-truth 3D curve point sets. These metrics are representation-agnostic and can be applied to both our method and the baselines.

\subsection{Results}

\begin{figure}[t]
  \centering
  \includegraphics[trim={200pt 80pt 200pt 20pt}, width=0.8\linewidth]{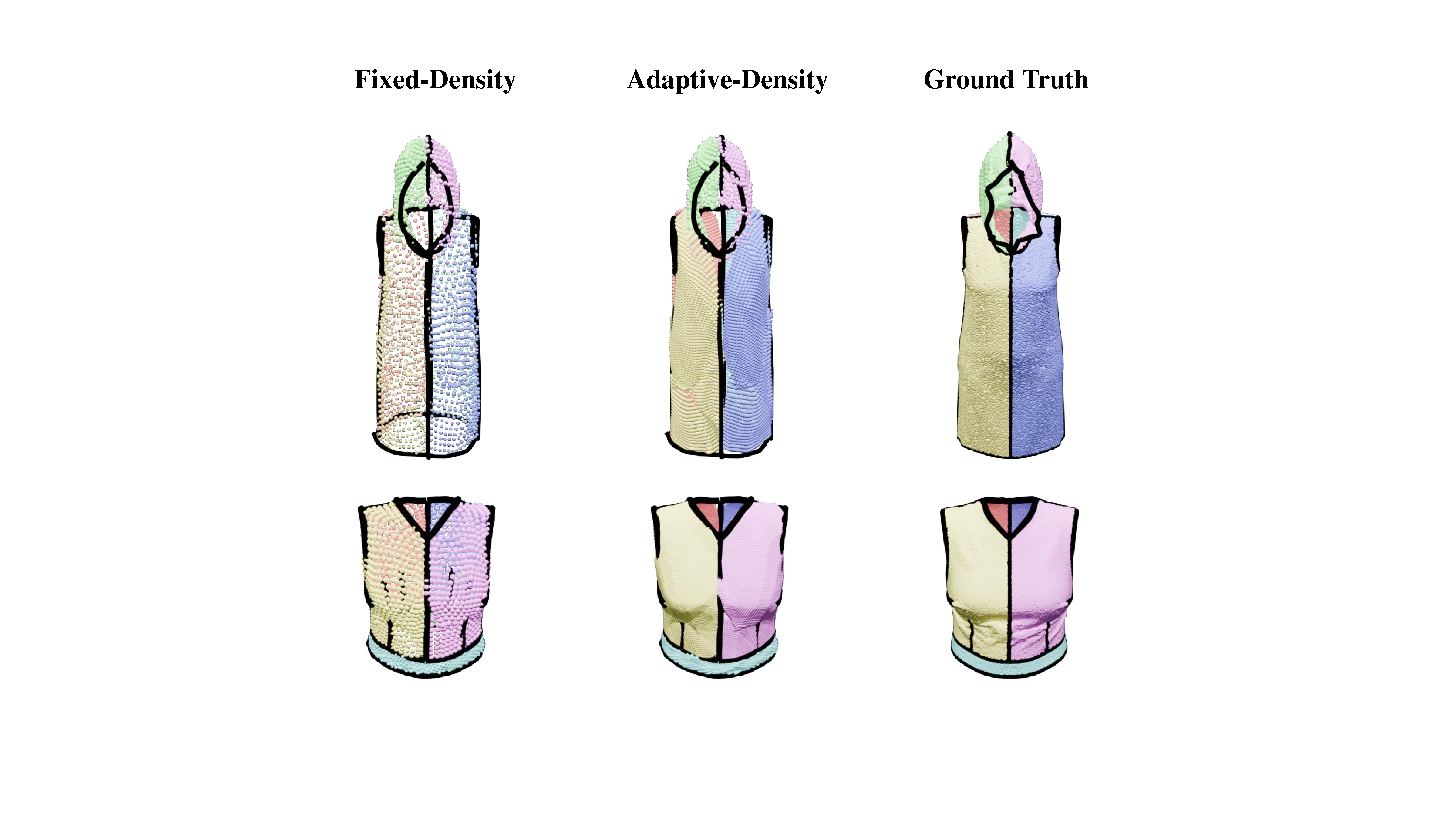}
  \caption{\textbf{Visualization on adaptive sampling density at inference time.} From left to right, the point set sampled at a fixed density of $20\times20$ points per patch; the point set sampled at the adaptive-density; the ground truth point cloud.}
  \label{fig:point_cloud_compare}
\end{figure}

\textbf{Geometry Reconstruction.} \Reweaver predicts patch geometries by mapping uniformly sampled coordinates from the unit square through hyper-networks. Although training uses a fixed $20\times20$ sampling density, the smoothness of the implicit mapping allows arbitrary sampling densities at inference. Because patch sizes vary significantly, a fixed sampling rate can make small patches appear overly dense and large patches overly sparse. To address this, we pre-sample a $20\times20$ grid and adaptively retain points based on spatial variance, producing a near-uniform point density across patches. While \citet{garmagenet} also predict 3D point clouds, its resolution is fixed and cannot adapt in this manner. Figure~\ref{fig:point_cloud_compare} compares our adaptive sampling with the ground-truth point clouds, and Table~\ref{tab:ablation_study} reports the corresponding CD scores. For curves, we use a fixed budget of 50 samples per curve, which is sufficient to capture all curve lengths without noticeable visual artifacts.

\textbf{2D Sewing Pattern Reconstruction.} We choose Sewformer~\cite{sewformer}, ChatGarment~\cite{chatgarment}, and AIpparel~\cite{aipparel} as our baseline models. For fairness, we replace their monocular image input with a four-view image input. 
Since both GCD-TS and the GCD-MM used by AIpparel for training essentially follow the same GCD distribution, we fine-tune AIpparel on our GCD-TS dataset until the loss curve plateaus rather than training it from scratch. 
For the same reason, we fine-tune ChatGarment instead of training from scratch, but convert the JSON files of our GCD-TS dataset into the refined version following the method described in the original paper. 
For the training of Sewformer, we adopt the modifications introduced in ~\cite{aipparel}.

Although our task uses multi-view inputs, the garment reconstruction task is more challenging on our dataset because we employ tileable textures rather than GCD’s default textures, which contain strong seam cues (see Figure~\ref{fig:texture-difference}). A visual comparison is shown in Figure~\ref{fig:pattern_compare}, and quantitative results are reported in Table~\ref{tab:pattern_compare}. 

\begin{figure}[htbp]
    \centering
    \begin{tikzpicture}
    \node[anchor=south west,inner sep=0] (image) at (0,0){\includegraphics[width=1\linewidth]{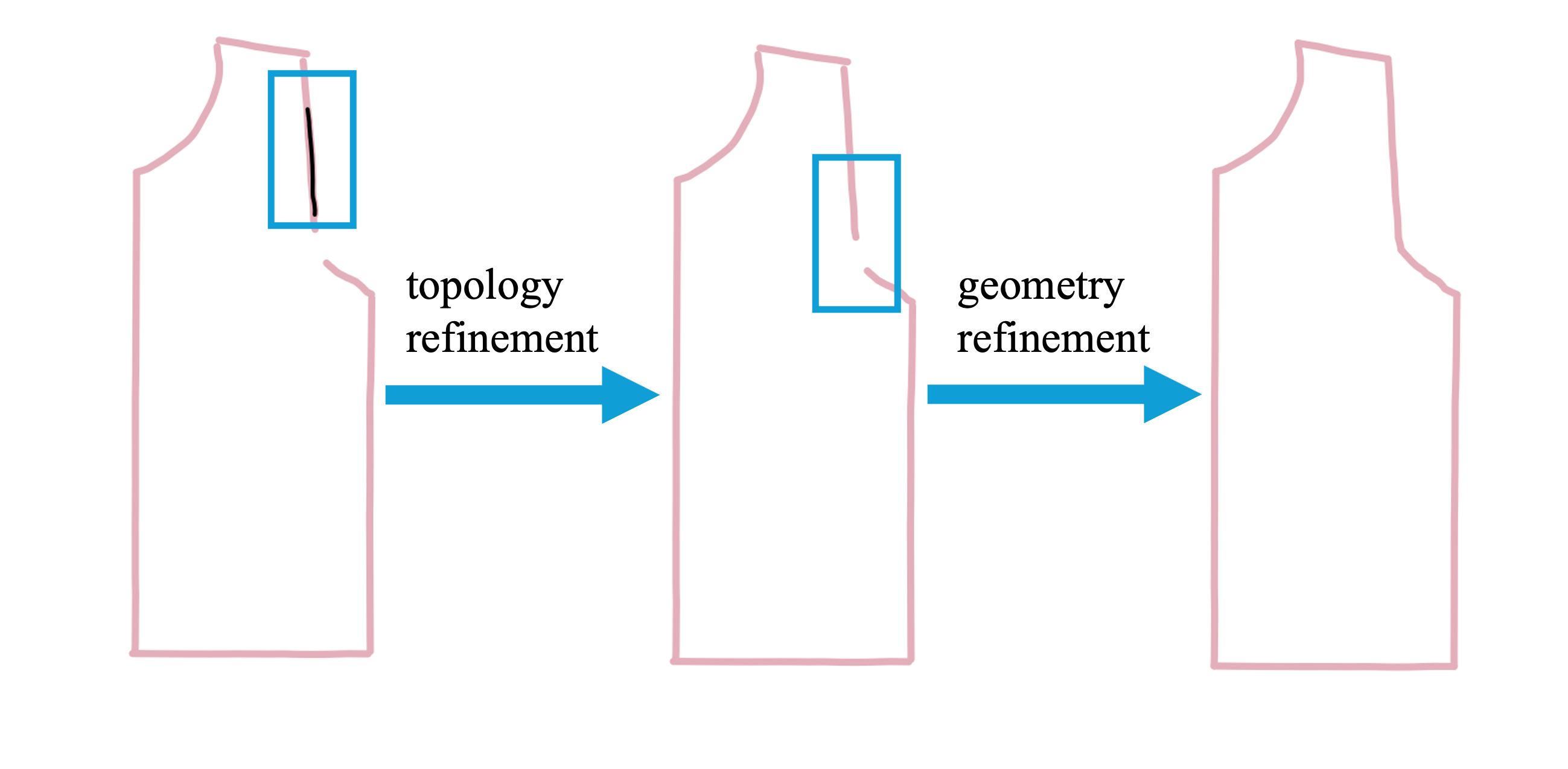}};

    \end{tikzpicture}
    \vspace{-0.5cm}
    \caption{\textbf{Topology and geometry refinement.} Without topology refinement, redundant 3D curves cause incorrect 2D edges (black segment in the leftmost image). Topology refinement removes these redundancies, producing clean panel structures, while geometry refinement enforces accurate, closed-loop boundaries suitable for triangulation.}
    \label{fig:ablation_study}
  \end{figure}

\begin{table}[htbp]
  \centering
  \caption{\textbf{Quantitative Comparison of 2D Panel Quality.} We compare the 2D panels generated by \Reweaver and baselines using the metrics defined in Section~\ref{subsec:exp:setup}. \Reweaver outperforms on five out of six metrics.}
  \vspace{-0.2cm}
  \footnotesize{
  \begin{tabular}{l|c|c|c|c|c}
    \toprule
        Method        
        & $\text{Acc}_p\uparrow$ 
        & $\text{Acc}_e\uparrow$ 
        & $\text{Acc}_o\uparrow$ 
        & $\text{CD}_e\downarrow$ 
        & IoU$\uparrow$  \\\hline

    Sewformer                &  0.3761   &  0.4802   &  0.1806  &  0.1161  &  0.5844\\
    
    ChatGarment             &  0.5557   &  \textbf{0.8012}   &  0.4452  &  0.0906  &  0.6533\\
    
    AIpparel & 0.4561 & 0.6774 & 0.3090 &0.0648  & 0.7084  \\

    \Reweaver  & \textbf{0.9210} & 0.7175 & \textbf{0.6608} & \textbf{0.0391} & \textbf{0.8221} \\
    \bottomrule
  \end{tabular}
  }
  \label{tab:pattern_compare}
\end{table}
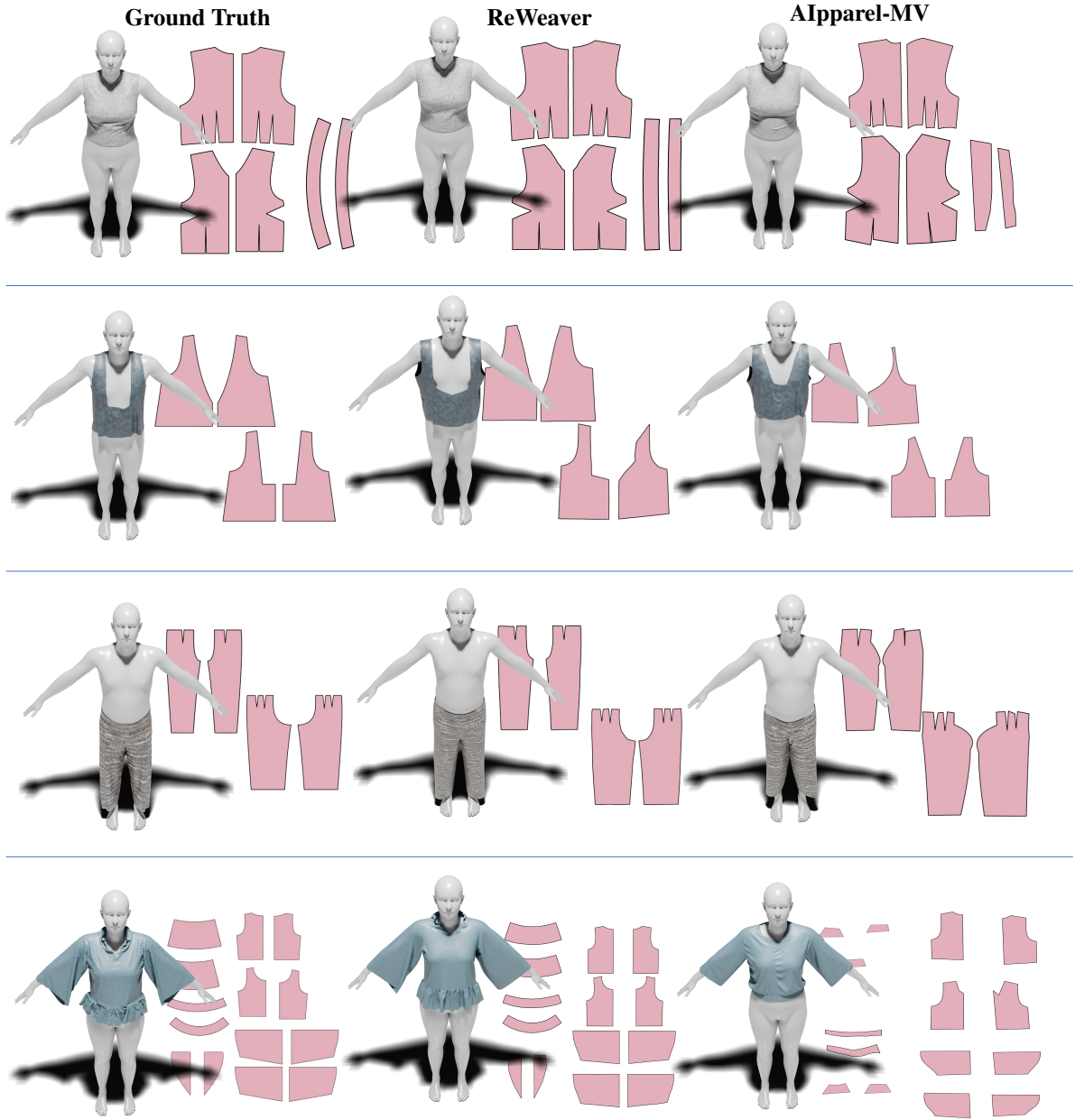
\begin{figure*}[htbp]
    \centering
    \begin{tikzpicture}
    \node[anchor=south west,inner sep=0] (image) at (0,0){\includegraphics[width=0.88\linewidth]{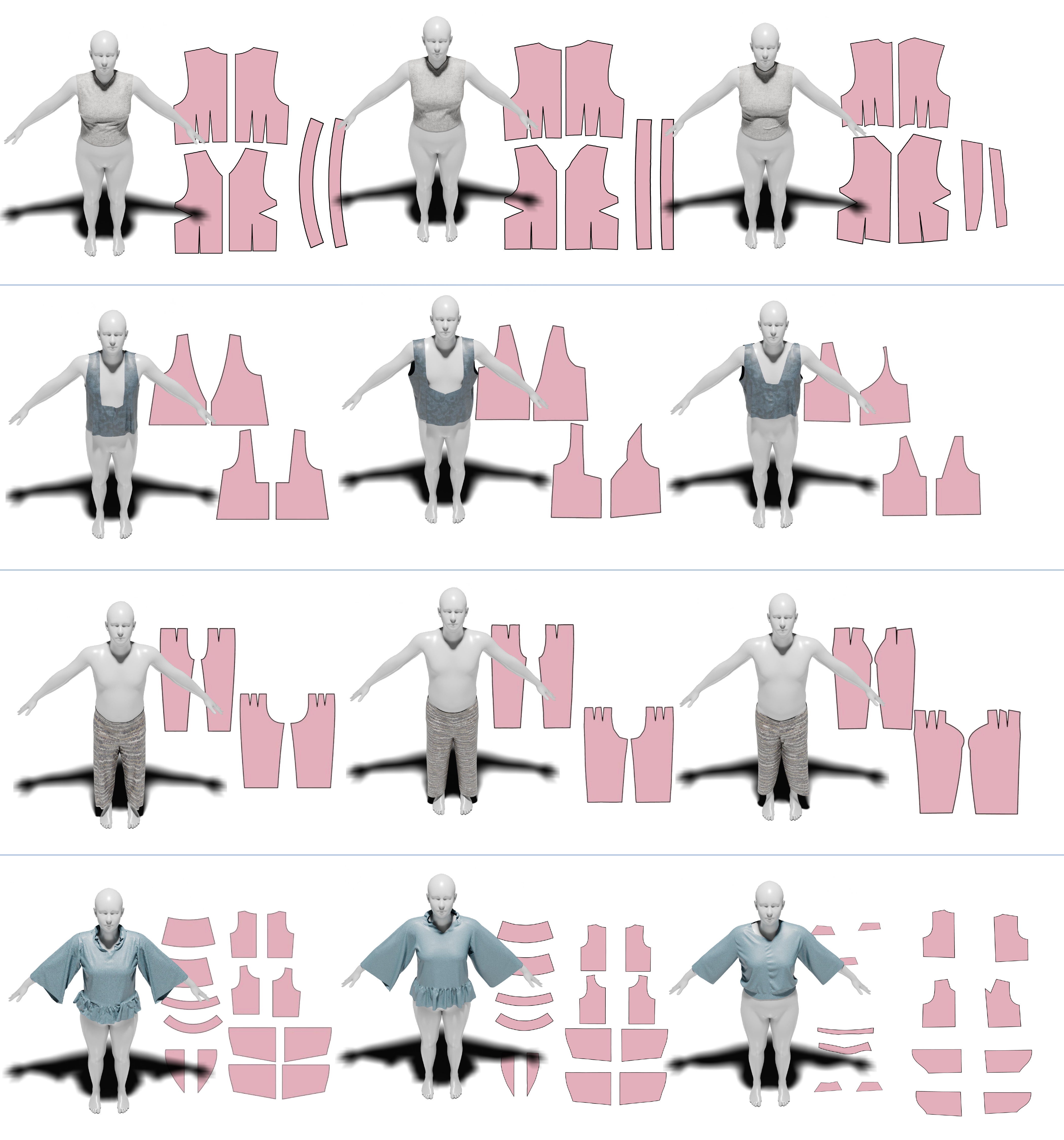}};
    \begin{scope}[x={(image.south east)},y={(image.north west)}]
        \node[anchor=south] at (0.18, 0.97) {\textbf{Ground Truth}};
        \node[anchor=south] at (0.5, 0.97) {\textbf{ReWeaver}};
        \node[anchor=south] at (0.8, 0.97) {\textbf{AIpparel}};
    \end{scope}
    \end{tikzpicture}
    \vspace{-0.5cm}
    \caption{\textbf{Comparison with AIpparel.} We compare \Reweaver and \textbf{AIpparel} against ground truth. Each example shows the predicted and ground-truth 2D panels along with the resulting simulated meshes. \Reweaver yields more accurate panels and correspondingly improved simulation results.}
    \label{fig:pattern_compare}
  \end{figure*}
\begin{table*}[htbp]
  \centering
  \caption{\textbf{Effects of topology and geometry refinement.} Refinement removes redundant edges and enforces closed boundaries, yielding higher Edge Acc and improved geometric quality (lower Edge CD and higher Panel IoU).}
  \small
  {
  \vspace{-0.2cm}
  \label{tab:ablation_study}
  \begin{tabular}{l|c|c|c|c|c|c|c|c}
    \toprule
       & $\text{CD}_p^{\text{base}}\downarrow$ 
       & $\text{CD}_p^{\text{adapt}}\downarrow$ 
       & $\text{CD}_c\downarrow$ 
       & $\text{Acc}_p\uparrow$ 
       & $\text{Acc}_e\uparrow$ 
       & $\text{Acc}_o\uparrow$ 
       & $\text{CD}_e\downarrow$ 
       & IoU$\uparrow$ \\\hline

    with refine-correction  &0.0232 & \textbf{0.0185} & 0.0266 & \textbf{0.9210} & \textbf{0.7175} & \textbf{0.6608} & \textbf{0.0391} & \textbf{0.8221} \\
        
    without refine   & \textbf{0.0225} & 0.0188 & \textbf{0.0255} & 0.9101 & 0.5361 & 0.4880 & 0.0416 & 0.7775 \\
    \bottomrule
  \end{tabular}
  }
\end{table*}

\textbf{Ablation Study.} We evaluate the impact of the topology and geometry refinement procedures described in Appendix~\ref{app:topo:refine} and~\ref{app:geo:refine}. Topology refinement removes duplicated or redundant edges, yielding a substantial improvement in Edge Count Accuracy ($\text{Acc}_e$) and enabling more accurate 2D panel reconstruction. Geometry refinement closes small gaps between edges in 2D space, producing fully closed panel boundaries. Figure~\ref{fig:ablation_study} shows qualitative differences, and Table~\ref{tab:ablation_study} reports the corresponding quantitative results.


\section{Conclusion}
\label{sec:conclusion}

We presented \Reweaver, an integrated framework for multi-view garment reconstruction that supports both perception and pattern recovery. \Reweaver produces 3D geometries that align closely with sparse-view inputs and reconstructs accurate 2D sewing patterns suitable for downstream physical simulation. Our experiments demonstrate that jointly reasoning over geometric primitives in both 2D and 3D spaces leads to notable improvements in garment topology and geometric fidelity.

{
\clearpage
    \small
    \bibliographystyle{ieeenat_fullname}
    \bibliography{main}
}

\appendix
\clearpage
\setcounter{page}{1}
\maketitlesupplementary

\section{Topology Refinement}
\label{app:topo:refine}
To obtain clean sewing-structure topology from the raw predictions, we apply a refinement pipeline that combines reliability-based filtering, 3D curve consolidation, and 2D loop analysis. This removes redundant or inconsistent curves and recovers stable, closed panel boundaries.

We begin by filtering patch predictions using a relatively high threshold $\epsilon_p = 0.7$. Since patch validity is already highly reliable ($\text{Acc}_p = 0.9210$), this single cutoff cleanly separates valid from invalid patch queries.

Curve validity and patch--curve adjacencies, in contrast, are filtered with more permissive thresholds, $\epsilon_c = 0.5$ and $\epsilon_{\text{adj}} = 0.5$. Empirically, these thresholds avoid false negatives but may retain redundant or overlapping curve predictions. The following refinement steps address this over-retention.

\paragraph{Duplicate Curve Merge}
We first merge curves that represent nearly the identical geometry. Two curves $C_a$ and $C_b$ are considered duplicates if their bidirectional Chamfer distance is small, i.e.,
\begin{equation*}
    \text{CD}(C_a \to C_b) < 0.03 \quad \text{and} \quad
\text{CD}(C_b \to C_a) < 0.03 .
\end{equation*}
For each such pair, we retain the curve with the higher predicted validity probability and transfer the discarded curve's adjacency relations to it.

\paragraph{Sub-curve Removal} 
If two curves share the same adjacency pattern, and one is geometrically contained inside the other, we treat the shorter one as a spurious ``sub-curve.'' Formally, if
$$
\text{CD}(C_{\text{sub}} \rightarrow C_{\text{main}}) < 0.04 ,
$$
we discard $C_{\text{sub}}$. This removes small fragments arising from local over-segmentation.

\paragraph{2D loop pruning}
For each panel, let $\{E_{ij}\}_{j=1}^N$ denote the incident 2D edges predicted from the associated 3D curves. Our goal is to keep only the set of edges that best forms a closed loop.

We first consider all possible orderings and orientations of the edges to form a closed boundary:
\[
[E'_{ij_k}] = (E'_{ij_1},\, E'_{ij_2},\,\ldots, E'_{ij_N}).
\]
For any ordering, we define the loop cost as
\begin{equation}
C([E'_{ij_k}]) 
= \sum_{k=1}^{N}
\left\|
E'_{ij_k}[-1] \;-\; E'_{ij_{k+1}}[0]
\right\|_2^2 ,
\qquad
j_{N+1} \!=\! j_1 .
\label{eq:loop-cost}
\end{equation}
The optimal cost for the unordered edge set is then
\begin{equation}
C(\{E_{ij}\}_{j=1}^N)
=\min_{[E'_{ij_k}]} C([E'_{ij_k}]).
\end{equation}

For any edge $E_{ik}$ in the panel, we compute the loop cost after removing it:
\[
C\Bigl(\{E_{ij}\}_{j=1}^N \setminus \{E_{ik}\}\Bigr).
\]
If removal reduces the optimal cost by a sufficient margin,
\begin{equation}
C\Bigl(\{E_{ij}\}_{j=1}^N \setminus \{E_{ik}\}\Bigr)
\;<\;
C\Bigl(\{E_{ij}\}_{j=1}^N\Bigr) - \epsilon,
\end{equation}
where $\epsilon = 10^{-9}$, then $E_{ik}$ is deemed inconsistent with the panel loop and is pruned. When multiple edges satisfy the condition, we remove the one that yields the greatest cost reduction. The pruning is repeated until convergence.

\subsection*{Ablation Analysis}
To isolate the effect of each refinement step, we conduct a cumulative ablation study in which the following rules are progressively enabled:
\begin{enumerate}
\item threshold-based filtering only;
\item thresholding + 2D loop pruning;
\item thresholding + loop pruning + sub-curve removal;
\item thresholding + loop pruning + sub-curve removal + duplicate merging (full refinement).
\end{enumerate}
The results in Table~\ref{tab:more_ablation_study} show that each refinement stage contributes positively to edge accuracy and panel IoU. Because topology refinement has negligible influence on the Chamfer Distance of patches and curves, we omit those metrics here.

\begin{table}[t]
  \centering
  \small
  \begin{minipage}{\linewidth}
    \centering
    \caption{\textbf{More ablation on topology refinement.} Progressive rules further improve edge accuracy and panel IoU and slightly reduce edge Chamfer Distance $\text{CD}_e$.}
    \vspace{-0.2cm}
    \label{tab:more_ablation_study}
    \begin{tabular}{lccc}
      \toprule
      Method 
      & $\text{Acc}_e\uparrow$ 
      & $\text{CD}_e \downarrow$
      & IoU$\uparrow$ \\
      \midrule
      Threshold only    
      & 0.5361  & 0.0416 & 0.7775  \\
      + 2D loop pruning
      & 0.6045 & 0.0444 & 0.7584 \\
      + sub-curve removal
      & 0.6269 & 0.0418 & 0.7940 \\
      + duplicate merging
      & \textbf{0.7175} & \textbf{0.0391} & \textbf{0.8221} \\
      \bottomrule
    \end{tabular}
  \end{minipage}
\end{table}

\section{2D Geometry Refinement}
\label{app:geo:refine}
Since our model does not explicitly enforce perfect edge-to-edge connectivity, the predicted panel boundaries may contain small gaps and misalignments. To obtain visually smoother and more aesthetically pleasing panels, we apply a geometry refinement stage. We first reorder and flip the edges as described in Section~\ref{app:topo:refine}, and then perform the following two geometry refinement steps:

\paragraph{Replace bad edges.}
Consider an ordered closed loop of $N$ edges $\{E_{ij}\}$ for panel $i$, where each edge $E_{ij} \in \mathbb{R}^{M_E \times 2}$ is represented by $M_E$ sampled 2D points. For each edge $E_{ij}$, we denote $s_j$ and $e_j$ as the start and end point, and similarly $e_{j-1}$ and $s_{j+1}$ from its neighbors (indices modulo $N$). 
We define the local connection gaps as:

\begin{equation*}
\text{gap}_1(j) = \bigl\| s_j - e_{j-1} \bigr\|_2,\quad
\text{gap}_2(j) = \bigl\| e_j - s_{j+1} \bigr\|_2,
\end{equation*}

and normalize them by the edge length $\ell_j = \|e_j - s_j\|_2$. Edges with a large normalized connection error,
\begin{equation}
\frac{\text{gap}_1(j)}{\ell_j} + \frac{\text{gap}_2(j)}{\ell_j} > \tau_{\text{gap}},
\end{equation}
are marked as "bad", where $\tau_{\text{gap}} = 3.0$. For each bad edge $E_{ij}$, we discard its original geometry and replace it with a straight segment that linearly interpolates between $e_{i-1}$ and $s_{i+1}$, resampled into $M_E$ points. This operation preserves the loop topology while smoothing out locally inconsistent or noisy edge predictions.

\paragraph{Align edges to joint midpoints.}
After replacing obviously bad edges, we further refine the geometry by snapping all edges to a consistent set of joint targets along the loop. For a panel $i$ with an ordered closed loop of edges $\{E_{ij}\}_{j=1}^N$, each sampled as $E_{ij} \in \mathbb{R}^{M_E \times 2}$, we let $e_j$ and $s_j$ denote the end point and start point of edge $E_{ij}$, respectively. For the joint between $E_{ij}$ and $E_{i,\,j+1}$ (indices taken modulo $N$), we define a target vertex
\begin{equation}
v_j = \frac{e_j + s_{j+1}}{2},
\end{equation}
i.e., the midpoint of the two incident edge endpoints. This yields a set of $N$ joint targets $\{v_j\}_{j=1}^N$ distributed along the loop.

For each edge $E_{ij}$, we estimate a 2D similarity transform
\begin{equation*}
    T_{ij}(p) = s_{ij} R_{ij} p + t_{ij},
\end{equation*}
that aligns its endpoints to the corresponding joint targets:
\begin{equation*}
    T_{ij}(s_j) = v_{j-1}, \qquad T_{ij}(e_j) = v_j.
\end{equation*}
Here $s_{ij} \in \mathbb{R}$ denotes the scale factor, $R_{ij} \in \mathbb{R}^{2 \times 2}$ is a 2D rotation matrix, and $t_{ij} \in \mathbb{R}^2$ is the translation vector. Such a solution clearly exists and is unique in the two-dimensional space. The scale parameter is clamped to a reasonable range to avoid degenerate transforms. After solving for $T_{ij}$, we apply it to all $M_E$ sampled points of $E_{ij}$ to obtain the refined edge $\tilde{E}_{ij}$. This alignment step enforces consistent edge-to-edge joints, removes small gaps or overlaps, and preserves the overall shape of each edge.

\section{Experiment Details} 
\label{app:exp:details}

\begin{figure*}[h]
    \centering
    \begin{tikzpicture}
    \node[anchor=south west,inner sep=0] (image) at (0,0){\includegraphics[width=\linewidth]{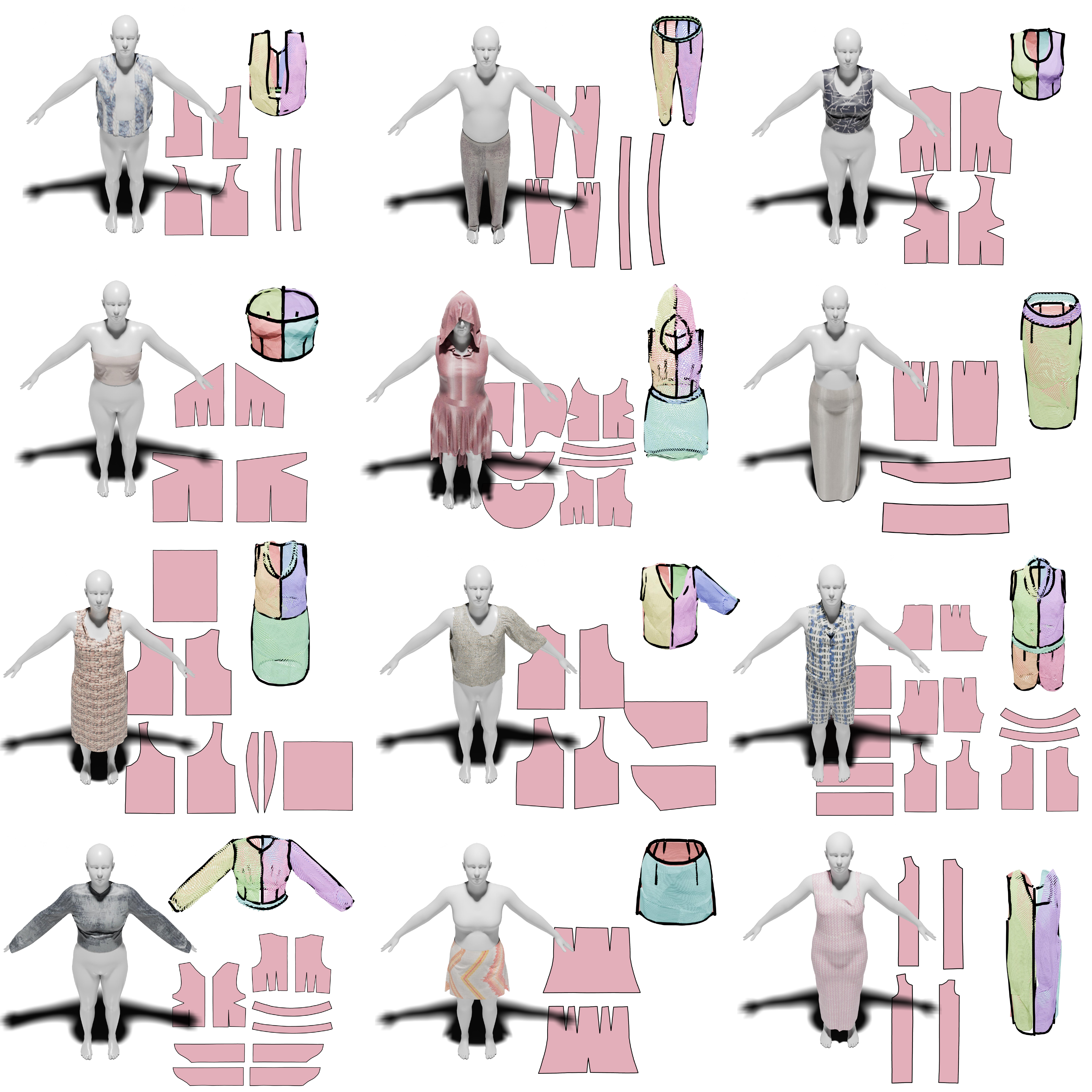}};
    \end{tikzpicture}
    \caption{Visualization of more results}
    \label{fig:more_result_in_app}
  \end{figure*}

\paragraph{More training  details.}
We train our model for 50 epochs on 8$\times$H200 GPUs, which takes about 120 hours in total. We use the AdamW optimizer with a unified weight decay of $1\times 10^{-4}$, and set the base learning rates of the visual encoder, 3D prediction head, and 2D prediction head to $5\times 10^{-5}$, $1\times 10^{-5}$, and $5\times 10^{-5}$, respectively. All three modules share the same iteration-wise schedule: a short linear warmup of 1 epochs that ramps the learning rate from a very small value to its base value, followed by cosine annealing down to a small floor of $5\times 10^{-6}$. We use a total batch size of 128 (16 per GPU across 8 GPUs).

\paragraph{Visual Encoder.} We adopt a lightweight DINO-based image encoder with 22M parameters, followed by a stack of 12 intra-frame and inter-frame attention layers. We use the DINOv2\cite{dinov2} backbone with a patch size of 14, and its weights are kept fully trainable during training. Inside the visual encoder, the token dimension is fixed to 384. The outputs of the last intra-frame attention layer and the last inter-frame attention layer are concatenated to form a 768-dimensional feature, which is used as the input to the subsequent modules. We resize all input images to a resolution of $518\times 518$ and apply per-pixel normalization by subtracting the dataset mean and dividing by the dataset standard deviation.

\paragraph{3D Curve and Patch Prediction.} We set the numbers of curve and patch queries to 200 and 70, respectively—about twice the maximum values in the dataset—to provide a safe margin for more complex or augmented cases while keeping the number of empty queries and the computational cost reasonably low. We assign a weight of 1 to all BCE losses and a weight of 300 to all geometric losses. We train curve/patch validity with a weighted binary cross-entropy loss: the positive class (valid curve/patch) has weight 1, and the negative class (empty query/patch) weight is chosen from the empirical ratio between valid curves and empty queries, scaled by a global factor so that both contribute roughly equally to the loss. We use a bipath transformer decoder with 12 layers, where each layer applies 1) per-path self-attention; 2) cross-attention between the two paths conditioned on primitive-type embeddings; 3) cross-attention to the image features; and 4) a final feed-forward network.

\paragraph{2D Pattern Prediction.} We assign a weight of 300 to the edge geometry loss and $1\times 10^{-2}$ to the scale loss. Each edge is represented by 50 points that are uniformly sampled along its arc length. For every panel, we normalize its 2D coordinates to the range $[-1, 1]$ by subtracting the panel-wise mean and dividing by the maximum absolute coordinate value (scale). This normalization improves the numerical stability of the predictions and largely removes the ambiguity caused by global translations.
The 2D pattern prediction module takes curve and patch features together with the patch–curve adjacency matrix, and first groups the corresponding tokens into panel-wise sets according to their adjacency relations. The panel-wise edge tokens are then processed by 12 stacked transformer-style layers: each layer applies self-attention among edges of the same panel, followed by cross-attention where edge tokens attend to the corresponding panel token (edges as queries, panel token as keys/values), and finally a feed-forward network update, yielding the edge geometry and a scalar scale prediction for each panel.

\section{More Results}
\label{app:more:results}
The results of our experiments on various styles, such as strapless dresses, pencil skirts, and asymmetrical tops, are shown in Figure~\ref{fig:more_result_in_app}.

\section{Garment Simulation}
\label{app:more:garment_simulation}
To dress the target human model, 2D garment panels are first positioned in 3D space based on their corresponding point cloud data. These panels are then imported into Marvelous Designer~\cite{md_web} along with the body model, where their placement is manually refined to ensure alignment with the respective body regions. 
Furthermore, the sewing relationships are established using the predictions from our model, which identifies and stitches together the edges corresponding to the same 3D curves. 
Subsequently, we employ the software’s built-in physics engine with default material parameters to simulate the draping process. This wraps the panels around the body under collision constraints, achieving a final dressed state that conforms to the target pose. The associated project files will be released in our code repository.

\section{Discussion}
\label{app:discussion}
Notably, compared to prior methods that predict structured pattern descriptions including seam connectivity~\cite{aipparel,dresscode}, our approach is inherently more flexible and potentially more generalizable to topology. Since we operate directly on panels and their 2D edge geometry rather than committing to a fixed Structured pattern representation, our method can in principle accommodate garments with additional components (e.g., sewn-on pockets) or complex topologies where seams are not strictly one-to-one at the edge level, without any architectural changes. A thorough empirical study of this generalization capability, especially on such challenging cases, is left for future work.

Unlike most previous works that focus on either 3D garment geometry or 2D panel layouts in isolation, our method reconstructs both representations jointly, which naturally enables further joint optimization to refine them simultaneously. Beyond this, one could extend our framework with additional prediction heads or differentiable cloth simulation to also estimate physical material parameters and human shape/pose from multi-view images, using the coupled 2D–3D predictions as constraints. This would allow reconstruction of more complete and better aligned 3D assets from real-world image, and we believe this constitutes a particularly interesting direction for future work.

\section{Limitations}
\label{app:limitations}
A key limitation lies in the scarcity of high-quality 3D garments with complex topology and photorealistic textures. 
This limits the model's generalization ability on out-of-distribution and real-world data. Future work could introduce greater variations in camera parameters, human shapes and poses, and backgrounds to generate more diverse synthetic data, and incorporate real-world data during training to narrow the sim-to-real gap.



\end{document}